\documentclass[12pt,journal]{IEEEtran}

\usepackage{amsmath,amssymb,amsthm}
\usepackage{graphicx}
\newcommand{\safeincludegraphics}[2][]{%
  \IfFileExists{#2}{\includegraphics[#1]{#2}}{%
    \fbox{\parbox[c][0.55\linewidth][c]{0.9\linewidth}{\centering Missing figure: \texttt{\detokenize{#2}}}}%
  }%
}
\usepackage{booktabs}
\usepackage{tabularx}
\usepackage{array}
\usepackage{multirow}
\usepackage{cite}
\usepackage[hidelinks]{hyperref}
\usepackage{xcolor}
\usepackage{bm}
\usepackage{algorithm}
\usepackage{algorithmic}
\usepackage{caption}
\usepackage{subcaption}

\newtheorem{theorem}{Theorem}
\newtheorem{proposition}{Proposition}
\newtheorem{corollary}{Corollary}
\newtheorem{definition}{Definition}
\newtheorem{remark}{Remark}

\newcommand{\R}{\mathbb{R}}
\newcommand{\Id}{\mathbf{I}}
\newcommand{\norm}[1]{\left\|#1\right\|}
\newcommand{\inner}[2]{\langle #1,\, #2 \rangle}

\begin{document}

\title{Exploring Oversmoothing with Householder Matrices}

\author{%
  Bhaskar Karol\\
  \textit{Preprint}\\
  \today
}

\maketitle

\begin{abstract}
Deep graph neural networks (GNNs) suffer from oversmoothing---a progressive collapse
of node representations toward a low-information subspace as network depth
increases---because the normalized graph propagation operator is repeatedly applied directly to the hidden representations. In this work we study \textbf{Householder Graph Neural Network
(HouseGNN)}. Rather than updating the
hidden state like standard GCN, HouseGNN uses the aggregated neighbourhood message solely to
estimate a reflection direction; the node embedding is then updated by a
Householder reflector followed by GroupSort, yielding a \emph{piecewise
orthogonal} layer that preserves Euclidean norm at every node and at
every depth.  We prove three core properties: (i) every internal layer preserves the node-wise Euclidean norm; (ii) the Householder
reflector is scale and sign-invariant in the message; and (iii) pairwise
distances between nodes can change through mismatch between node-wise orthogonal operators. 
\end{abstract}

\begin{IEEEkeywords}
Graph Neural Networks, Oversmoothing, Householder Reflection, Orthogonal Networks,
GroupSort.
\end{IEEEkeywords}

\section{Introduction}
\label{sec:intro}

Graph neural networks (GNNs)~\cite{kipf2017semi,velickovic2018gat,xu2019how}
have become the dominant area for learning on graph-structured data, achieving
strong empirical performance in node classification~\cite{wu2019simplifying},
link prediction, and molecular property prediction~\cite{duvenaud2015convolutional}.
The majority of effective architectures follow the \emph{message-passing} framework
\cite{gilmer2017neural}: at each layer, every node aggregates features from its
neighbours and updates its representation via a learnable transformation.

Despite strong shallow performance, deep GNNs are well-known to suffer from
\emph{oversmoothing}~\cite{li2018deeper,oono2020graph}. As depth grows, repeated
application of the graph propagation operator suppresses high-frequency spectral
components, progressively reducing discriminative variation across nodes and
driving the representation toward a low-dimensional subspace determined by the
leading eigenspace of the propagation operator~\cite{oono2020graph}. In node
classification, this loss of variation can make node representations increasingly
difficult to separate.

A variety of strategies have been proposed to combat oversmoothing, including
PairNorm~\cite{zhao2020pairnorm}, GCNII~\cite{chen2020simple},
DropEdge~\cite{rong2020dropedge}, residual connections~\cite{he2016deep},
and batch normalisation~\cite{ioffe2015batch}.

In this work, we study oversmoothing using Householder matrices. Instead of directly updating the hidden states, we change \emph{how the hidden state is updated}.  The key idea is to use the neighbourhood message
only to define a \emph{reflection direction}, then update the node state by
a Householder reflection~\cite{householder1958unitary} across the hyperplane
normal to that direction, followed by GroupSort~\cite{anil2019sorting}, a
norm-preserving nonlinearity.

\section{Related Work}
\label{sec:related}

\subsection{Oversmoothing in GNNs}
The oversmoothing phenomenon was first identified by Li~et~al.~\cite{li2018deeper},
who observed empirically that GCN performance degrades with depth.  Oono and
Suzuki~\cite{oono2020graph} later proved that GCN representations
converge to a subspace at an exponential rate.  Wu~et~al.~\cite{wu2023nonasymptotic}
provided a non-asymptotic analysis showing that oversmoothing can appear even in
shallow models under specific random graph conditions.  Oversmoothing in
attention-based GNNs was studied by Wu~et~al.~\cite{wu2023demystifying}.

\subsection{Mitigation Strategies}
A number of architectural modifications reduce oversmoothing.  Zhao and
Akoglu~\cite{zhao2020pairnorm} propose PairNorm, which rescales representations
to maintain a fixed total pairwise distance.  Chen~et~al.~\cite{chen2020simple}
introduce GCNII, which combines initial residual connections with identity mappings
at each layer.  Rong~et~al.~\cite{rong2020dropedge} randomly drop edges during
training to limit graph smoothing.  Residual GCN~\cite{he2016deep} and
APPNP~\cite{gasteiger2019predict} retain initial-feature information through
skip connections.  Jumping Knowledge Networks~\cite{xu2018representation} aggregate
representations from all intermediate layers.  Scholkemper~et~al.~\cite{scholkemper2025residual}
prove that residual connections and normalisation can provably prevent
oversmoothing under certain conditions.

\subsection{Orthogonal Parameterisations in GNNs}
Several recent works use orthogonality constraints in GNNs.  Guo~et~al.~\cite{guo2021orthogonal}
propose OOGNN with orthogonal weight matrices to prevent oversmoothing.
Kiani~et~al.~\cite{kiani2024unitary} study unitary convolutions by modifying the Adjacency matrix.

\subsection{Householder Transformations}
Householder reflectors~\cite{householder1958unitary} are used for orthogonal matrix factorisation (QR
decomposition) and eigenvalue computation.  Their use as learnable
parameterisations in neural networks was explored by Mhammedi~et~al.~\cite{mhammedi2017efficient}
in the context of recurrent networks requiring fast orthogonal transformations.
In this work, Householder reflectors serve a different purpose: rather than
parameterising a global weight matrix, each reflector is \emph{conditioned
dynamically} on the node's neighbourhood, making it a local, graph-dependent,
per-node transformation.

\section{Background}
\label{sec:background}

\begin{table}[!t]
\centering
\caption{Notation used in the work.}
\label{tab:notation}
\renewcommand{\arraystretch}{1.1}
\setlength{\tabcolsep}{4pt}
\footnotesize
\begin{tabularx}{\columnwidth}{@{}p{0.30\columnwidth}X@{}}
\toprule
\textbf{Symbol} & \textbf{Meaning} \\
\midrule
$\mathcal{G}=(\mathcal{V},\mathcal{E})$ & Undirected graph with node set $\mathcal{V}$ and edge set $\mathcal{E}$ \\
$n=|\mathcal{V}|$ & Number of nodes \\
$\mathbf{x}_i\in\R^{d_{\mathrm{in}}}$ & Input feature vector of node $i$ \\
$X\in\R^{n\times d_{\mathrm{in}}}$ & Stacked input feature matrix \\
$\mathbf{h}_i^{(\ell)}\in\R^d$ & Hidden state of node $i$ at layer $\ell$ \\
$H^{(\ell)}\in\R^{n\times d}$ & Stacked hidden-state matrix \\
$A$ & Adjacency matrix of the graph \\
$\widetilde{A}=A+I_n$ & Adjacency matrix with self-loops \\
$\widetilde{D}$ & Degree matrix of $\widetilde{A}$ \\
$S=\widetilde{D}^{-1/2}\widetilde{A}\widetilde{D}^{-1/2}$ & Symmetric normalized adjacency used in GCN \\
$P=\widetilde{D}^{-1}\widetilde{A}$ & Row-normalized adjacency / mean-aggregation matrix \\
$\mathbf{m}_i^{(\ell)}$ & Mean neighborhood message of node $i$ \\
$\mathbf{u}_i^{(\ell)}$ & Unit reflection direction of node $i$ \\
$R_i^{(\ell)}$ & Householder reflector at node $i$ \\
$\Pi_i^{(\ell)}$ & Local permutation induced by GroupSort \\
$O_i^{(\ell)}=\Pi_i^{(\ell)}R_i^{(\ell)}$ & Full node-wise update operator \\
\bottomrule
\end{tabularx}
\end{table}

\subsection{Graph Neural Networks and Notation}

Let $\mathcal{G}=(\mathcal{V},\mathcal{E})$ be an undirected graph with
$n=|\mathcal{V}|$ nodes. Each node $i\in\mathcal{V}$ carries an input feature
vector $\mathbf{x}_i\in\R^{d_{\mathrm{in}}}$.

We use column-vector notation for per-node states throughout the work. Thus,
$\mathbf{h}^{(\ell)}_i\in\R^d$ and $\mathbf{m}^{(\ell)}_i\in\R^d$ are column
vectors. When the stacked matrix $H^{(\ell)}\in\R^{n\times d}$ is used, its
$i$-th row is $(\mathbf{h}^{(\ell)}_i)^\top$.

Let $A\in\R^{n\times n}$ denote the adjacency matrix of the graph, and let
\[
\widetilde{A}=A+I_n
\]
denote the adjacency matrix with self-loops. Its degree matrix is
\[
\widetilde{D}_{ii}=\sum_{j=1}^n \widetilde{A}_{ij}.
\]

Two normalized graph operators are used in this work. The first is the
row-normalized adjacency
\begin{equation}
  P=\widetilde{D}^{-1}\widetilde{A},
  \label{eq:row_normalized_operator_background}
\end{equation}
which is the random-walk or mean-aggregation operator. Its $i$-th row sums to
one, so it explicitly averages over the neighbourhood including the node itself.

The second is the symmetric normalized adjacency
\begin{equation}
  S=\widetilde{D}^{-1/2}\widetilde{A}\widetilde{D}^{-1/2},
  \label{eq:symmetric_normalized_operator_background}
\end{equation}
which is the propagation matrix used in the standard GCN layer. For undirected
graphs, $S$ is symmetric, whereas $P$ is generally not symmetric unless the graph
is regular.

These two are closely related. Since $\widetilde{D}$ is a
positive diagonal matrix after adding self-loops,
\begin{equation}
  S
  = \widetilde{D}^{1/2}P\widetilde{D}^{-1/2}.
  \label{eq:p_s_similarity_background}
\end{equation}
Thus $P$ and $S$ are similar matrices and therefore have the same eigenvalues.
They are not the same operator on node features, but they have the same spectral
convergence factors. This distinction is important for oversmoothing: both
operators have the same asymptotic smoothing rate on an undirected graph, while
their limiting representations differ by degree-dependent scaling.

\subsection{Standard Graph Convolutional Network}

The GCN update~\cite{kipf2017semi} at layer $\ell$ is
\begin{equation}
  H^{(\ell+1)} = \sigma\!\left(S\,H^{(\ell)}\,W^{(\ell)}\right),
  \label{eq:gcn}
\end{equation}
where $W^{(\ell)}\in\R^{d_\ell\times d_{\ell+1}}$ is a learnable weight
matrix and $\sigma$ is a nonlinearity.

\subsection{Spectral View of Oversmoothing}

Oversmoothing refers to the loss of node-wise discriminative variation caused by
repeatedly applying a graph propagation operator. In the linearized setting, where nonlinearities
and weight matrices are ignored, a GCN layer has this form
\begin{equation}
  H^{(\ell+1)} \approx M H^{(\ell)},
  \qquad
  H^{(L)} \approx M^L H^{(0)},
  \label{eq:linearized_smoothing}
\end{equation}
for a propagation operator $M$. Oversmoothing occurs when the powers $M^L$ suppress
all spectral components except those associated with the dominant eigenspace.
For the row-normalized operator $P=\widetilde{D}^{-1}\widetilde{A}$, each row
sums to one, hence
\begin{equation}
  P\mathbf{1}=\mathbf{1}.
\end{equation}
Thus $1$ is an eigenvalue of $P$. If the graph is undirected and connected, and
self-loops are included through $\widetilde{A}=A+I_n$, then $P$ is an irreducible
and aperiodic stochastic matrix. By the Perron--Frobenius theorem, one of the eigenvalue of $P$ is 
$1$ and every other eigenvalue satisfies
\begin{equation}
  |\lambda_k(P)|<1, \qquad k\ge 2.
  \label{eq:p_spectral_gap}
\end{equation}
Consequently,
\begin{equation}
  P^L \longrightarrow \mathbf{1}\pi^\top,
  \qquad
  \pi_i=\frac{\widetilde{D}_{ii}}{\sum_{j=1}^n \widetilde{D}_{jj}},
  \label{eq:p_limit}
\end{equation}
and therefore
\begin{equation}
  P^L H^{(0)} \longrightarrow \mathbf{1}\pi^\top H^{(0)}.
\end{equation}
In this row-normalized case, all node representations converge to the same vector.

For the symmetric GCN operator
\begin{equation}
  S=\widetilde{D}^{-1/2}\widetilde{A}\widetilde{D}^{-1/2},
\end{equation}
the same eigenvalue conclusion holds because $S$ is similar to $P$:
\begin{equation}
  S=\widetilde{D}^{1/2}P\widetilde{D}^{-1/2}.
  \label{eq:p_s_similarity_spectral}
\end{equation}
Similar matrices have the same eigenvalues, so $S$ also has only one eigenvalue
equals to $1$ and all other eigenvalues have modulus strictly less than $1$ under the
same connected, self-looped, undirected-graph assumption. Since $S$ is symmetric,
it admits an orthogonal eigendecomposition
\begin{equation}
  S=Q\Lambda Q^\top,
\end{equation}
and the linearized GCN propagation satisfies
\begin{equation}
  H^{(L)}\approx S^L H^{(0)} = Q\Lambda^LQ^\top H^{(0)}.
  \label{eq:spectral_gcn_power}
\end{equation}
Every component with $|\lambda_k|<1$ is multiplied by $\lambda_k^L$ and decays. The surviving eigenvector is proportional to
$\widetilde{D}^{1/2}\mathbf{1}$, so
\begin{equation}
  S^L \longrightarrow u_1u_1^\top,
  \qquad
  u_1=\frac{\widetilde{D}^{1/2}\mathbf{1}}
  {\sqrt{\mathbf{1}^\top\widetilde{D}\mathbf{1}}}.
  \label{eq:s_limit}
\end{equation}
Thus $S^L H^{(0)}$ converges to degree-scaled copies of a common feature vector:
node embeddings become parallel, although not necessarily identical.

Hence $P$ and $S$ describe the same spectral oversmoothing mechanism: they have
the same eigenvalues and therefore the same asymptotic spectral convergence.
The difference is in the limiting form. The row-normalized operator $P$ makes the
node embeddings identical, while the symmetric
operator $S$ makes them align in the same direction with degree-dependent
magnitudes. In both cases, repeated direct application of the graph operator to
the hidden state causes oversmoothing~\cite{oono2020graph}.

\subsection{Dirichlet Energy}

For a hidden-state matrix $H\in\mathbb{R}^{n\times d}$ with
row $i$ equal to $(\mathbf{h}_i)^\top$, define
\begin{equation}
  \mathcal{E}_{\mathrm{Dir}}(H)
  =
  \sum_{(i,j)\in \mathcal{E}_u}
  \norm{\mathbf{h}_i-\mathbf{h}_j}_2^2,
  \label{eq:dirichlet}
\end{equation}
where $\mathcal{E}_u$ denotes the set of unique undirected edges. Dirichlet
energy measures how much neighboring node representations vary across the
graph. Smaller values correspond to smoother representations over edges,
meaning that adjacent node representations tend to be more similar, whereas
larger values indicate greater variation between adjacent nodes. In particular,
on a connected graph, $\mathcal{E}_{\mathrm{Dir}}(H)=0$ if and only if all node
representations are identical.
\section{Householder Graph Neural Network}
\label{sec:method}

\subsection{Design Principle}
The central design decision is to \textbf{decouple neighbourhood
aggregation from state update}. The neighbourhood message is used only to define
a \emph{reflection direction}; the current node state is then reflected across
the hyperplane normal to that direction. Formally, for node $i$ at layer $\ell$,
the update is defined as follows.

\textbf{Step 1 --- Aggregate.} Compute the mean neighbourhood message:
\begin{equation}
  \mathbf{m}^{(\ell)}_i
  =
  \sum_{j=1}^{n} P_{ij}\,\mathbf{h}^{(\ell)}_j,
  \label{eq:aggregate}
\end{equation}
where $P\in\mathbb{R}^{n\times n}$ is the row-normalized mean-aggregation
operator induced by the graph.
\begin{equation}
  P=\widetilde{D}^{-1}\widetilde{A}
\end{equation}
Thus, \eqref{eq:aggregate} can equivalently be written as
\begin{equation}
  \mathbf{m}^{(\ell)}_i
  =
  \frac{1}{|\widetilde{\mathcal{N}}(i)|}
  \sum_{j\in \widetilde{\mathcal{N}}(i)}
  \mathbf{h}^{(\ell)}_j,
  \label{eq:mean_neighbourhood}
\end{equation}
where $\widetilde{\mathcal{N}}(i)=\mathcal{N}(i)\cup\{i\}$ denotes the
neighbourhood of node $i$ including itself.

\textbf{Step 2 --- Project direction.} Transform the message by an orthogonal weight
matrix and normalize:
\begin{equation}
  \mathbf{v}^{(\ell)}_i
  =
  W^{(\ell)}\mathbf{m}^{(\ell)}_i,
  \qquad
  (W^{(\ell)})^\top W^{(\ell)} = I_d.
  \label{eq:project}
\end{equation}
For $\mathbf{v}^{(\ell)}_i\neq \mathbf{0}$, define the unit reflection direction
as
\begin{equation}
  \mathbf{u}^{(\ell)}_i
  =
  \frac{\mathbf{v}^{(\ell)}_i}
  {\|\mathbf{v}^{(\ell)}_i\|_2}.
  \label{eq:normalise}
\end{equation}

\textbf{Step 3 --- Reflect.} Form the Householder reflector and apply it to the
current state:
\begin{equation}
  R^{(\ell)}_i
  =
  I_d
  -
  2\,\mathbf{u}^{(\ell)}_i
  (\mathbf{u}^{(\ell)}_i)^\top,
  \label{eq:reflector}
\end{equation}
\begin{equation}
  \mathbf{z}^{(\ell)}_i
  =
  R^{(\ell)}_i\,\mathbf{h}^{(\ell)}_i.
  \label{eq:reflect}
\end{equation}

\textbf{Step 4 --- Nonlinearity.} Apply GroupSort:
\begin{equation}
  \mathbf{h}^{(\ell+1)}_i
  =
  \mathrm{GroupSort}\!\left(\mathbf{z}^{(\ell)}_i\right).
  \label{eq:groupsort}
\end{equation}

\noindent
GroupSort~\cite{anil2019sorting} is a norm-preserving
nonlinearity. Let the group size be $g$, and assume that $d$ is divisible by $g$.
For a vector $\mathbf{z}\in\R^d$, write
\[
\mathbf{z}=
\big[\mathbf{z}^{(1)};\mathbf{z}^{(2)};\ldots;\mathbf{z}^{(d/g)}\big],
\]
where each block $\mathbf{z}^{(k)}\in\R^g$ contains $g$ consecutive coordinates.
GroupSort sorts the entries inside each block in ascending order and then
concatenates the sorted blocks.

In our experiments, we used group size $g=2$. For example, if
\[
\mathbf{z}=
\begin{bmatrix}
3\\
-1\\
2\\
5\\
4\\
4
\end{bmatrix},
\]
then
\[
\mathrm{GroupSort}(\mathbf{z})=
\begin{bmatrix}
-1\\
3\\
2\\
5\\
4\\
4
\end{bmatrix}.
\]

\subsection{Contrast with Standard GCN}

\begin{table}[!t]
\centering
\caption{Structural comparison between GCN and HouseGNN.}
\label{tab:compare}
\renewcommand{\arraystretch}{1.15}
\setlength{\tabcolsep}{3pt}
\footnotesize
\begin{tabularx}{\columnwidth}{@{}p{0.27\columnwidth}XX@{}}
\toprule
\textbf{Aspect} & \textbf{GCN} & \textbf{HouseGNN} \\
\midrule
Role of graph
& Forms new state
& Defines reflection direction \\

State update
& $\sigma(SHW)$
& $\mathrm{GroupSort}(R_i^{(\ell)}\mathbf{h}_i^{(\ell)})$ \\

Layer map
& Contractive / diffusive
& Piecewise orthogonal \\

Norm per layer
& Not preserved
& Preserved \\

\bottomrule
\end{tabularx}
\end{table}
The structural difference is that in GCN the graph operator acts
\emph{directly} on the hidden state, whereas in HouseGNN it uses a direction
vector $\{\mathbf{u}^{(\ell)}_i\}$ that governs local reflections.  Since message
magnitude is discarded in \eqref{eq:normalise}, adjacency eigenvalues can no
longer appear as repeated multiplicative shrinkage of the state.

\subsection{Geometric Interpretation}

For any unit vector $\mathbf{u}\in\R^d$, decompose a state
$\mathbf{h}\in\R^d$ as
\[
  \mathbf{h}
  = \underbrace{\inner{\mathbf{u}}{\mathbf{h}}\,\mathbf{u}}_{\text{along }\mathbf{u}}
  + \underbrace{(\mathbf{h} - \inner{\mathbf{u}}{\mathbf{h}}\,\mathbf{u})}_{\perp\,\mathbf{u}}.
\]
Then $R(\mathbf{u})\mathbf{h}
= -\inner{\mathbf{u}}{\mathbf{h}}\,\mathbf{u}
+ (\mathbf{h} - \inner{\mathbf{u}}{\mathbf{h}}\,\mathbf{u})$, i.e.\ the
reflector flips exactly the component along $\mathbf{u}$ and leaves the
orthogonal complement unchanged.  In HouseGNN, $\mathbf{u}$ is not fixed; it is
estimated from the node's neighbourhood.  The method therefore says \emph{``reflect
your current state across the hyperplane whose normal is estimated from your
neighbourhood''}, not \emph{``become the neighbourhood average''}.

\section{Theoretical Analysis}
\label{sec:theory}

All proofs are fully detailed in Section~\ref{sec:proofs}.

\subsection{Orthogonality and Spectral Structure}

\begin{proposition}[Orthogonality and eigenvalues]
\label{prop:ortho}
For any unit vector $\mathbf{u}\in\R^d$, the Householder matrix
$R(\mathbf{u})=\Id_d - 2\mathbf{u}\mathbf{u}^\top$ satisfies
\[
  R(\mathbf{u})^\top R(\mathbf{u}) = \Id_d,
  \qquad
  R(\mathbf{u})^\top = R(\mathbf{u}).
\]
Its spectrum is $\{-1\}$ (multiplicity 1, eigenvector $\mathbf{u}$) and
$\{+1\}$ (multiplicity $d-1$, eigenspace $\mathbf{u}^\perp$).
\end{proposition}

Proposition~\ref{prop:ortho} establishes that the reflector is both orthogonal
and an involution ($R^2=\Id$): it flips precisely one learned direction and
preserves all orthogonal directions.

\subsection{Layerwise Norm Preservation}

\begin{proposition}[Norm preservation]
\label{prop:norm}
Let $\mathbf{h}^{(\ell+1)}_i = \mathrm{GroupSort}(R^{(\ell)}_i \mathbf{h}^{(\ell)}_i)$.
Then $\norm{\mathbf{h}^{(\ell+1)}_i} = \norm{\mathbf{h}^{(\ell)}_i}$ for every
node $i$ and every layer $\ell$.  Consequently,
$\norm{\mathbf{h}^{(\ell)}_i} = \norm{\mathbf{h}^{(0)}_i}$ for all $\ell\ge 0$.
\end{proposition}

\subsection{Direction-Only Dependence}

\begin{proposition}[Scale and sign invariance]
\label{prop:scale}
Define $R(\mathbf{v}) = \Id_d - 2\frac{\mathbf{v}}{\norm{\mathbf{v}}}\frac{\mathbf{v}^\top}{\norm{\mathbf{v}}}$
for $\mathbf{v}\ne\mathbf{0}$.  Then for every nonzero scalar $c$,
\[
  R(c\mathbf{v}) = R(\mathbf{v}).
\]
\end{proposition}

The reflector depends only on $\mathrm{span}(\mathbf{v})$.  In vanilla GCN,
adjacency eigenvalues directly multiply the state repeatedly.  In HouseGNN, the
same propagation operator shapes the message $\mathbf{m}^{(\ell)}_i$, but message
magnitude is discarded in \eqref{eq:normalise} before any state update occurs.

\begin{definition}[Pairwise collapse]
Two nodes $i$ and $j$ are said to collapse at depth $L$ if
\[
\mathbf{h}_i^{(L)}=\mathbf{h}_j^{(L)}.
\]
More generally, a deep graph model is said to exhibit oversmoothing if, as depth
increases, hidden representations lose discriminative variation across many
nodes and become increasingly difficult to separate for node classification.
\end{definition}

\subsection{Obstruction to Pairwise Collapse}

\begin{proposition}[Equal-radius requirement for collapse]
\label{prop:collapse}
Suppose there exists a layer $L$ such that
$\mathbf{h}^{(L)}_i = \mathbf{h}^{(L)}_j$.
Then necessarily $\norm{\mathbf{h}^{(0)}_i} = \norm{\mathbf{h}^{(0)}_j}$.
\end{proposition}

\subsection{Pairwise Distance Evolution}

For each node $i$ and layer $\ell$, define the combined layer operator
\[
  O^{(\ell)}_i := \Pi^{(\ell)}_i R^{(\ell)}_i,
\]
where $\Pi^{(\ell)}_i$ is the local GroupSort permutation at node $i$.  Then
$\mathbf{h}^{(\ell+1)}_i = O^{(\ell)}_i\mathbf{h}^{(\ell)}_i$ and
$(O^{(\ell)}_i)^\top O^{(\ell)}_i = \Id_d$.  Define
\[
  \Delta^{(\ell)}_{ij} := \mathbf{h}^{(\ell)}_i - \mathbf{h}^{(\ell)}_j,
  \quad
  d^{(\ell)}_{ij} := \norm{\Delta^{(\ell)}_{ij}},
  \quad
  \rho := \max_k \norm{\mathbf{h}^{(0)}_k}.
\]

\begin{theorem}[One-step pairwise distance bound]
\label{thm:distance}
For any nodes $i,j$ and any layer $\ell$,
\[
  \left|d^{(\ell+1)}_{ij} - d^{(\ell)}_{ij}\right|
  \;\le\;
  \rho\,\left\|O^{(\ell)}_i - O^{(\ell)}_j\right\|_2.
\]
\end{theorem}

Theorem~\ref{thm:distance} identifies node-wise \emph{operator mismatch} as the
\emph{sole} mechanism by which pairwise distances can change.  The hidden state is
not averaged into a neighbourhood mean; if all nodes experience the same orthogonal
operator at layer $\ell$, then all pairwise distances are preserved exactly at that
layer.

\subsection{The Same-GroupSort Special Case}
\label{sec:special}

Define the absolute cosine similarity between message directions:
\begin{equation}
  \gamma^{(\ell)}_{ij}
  := \frac{|\inner{\mathbf{m}^{(\ell)}_i}{\mathbf{m}^{(\ell)}_j}|}
           {\norm{\mathbf{m}^{(\ell)}_i}\,\norm{\mathbf{m}^{(\ell)}_j}}
  \in [0,1].
  \label{eq:cosine}
\end{equation}
Because $W^{(\ell)}$ is orthogonal, $\gamma^{(\ell)}_{ij}$ equals the absolute
cosine similarity between the learned directions $\mathbf{u}^{(\ell)}_i$ and
$\mathbf{u}^{(\ell)}_j$.

\begin{proposition}[Exact reflector mismatch]
\label{prop:mismatch}
For any nodes $i,j$ and any layer $\ell$,
\[
  \left\|R^{(\ell)}_i - R^{(\ell)}_j\right\|_2
  = 2\sqrt{1 - (\gamma^{(\ell)}_{ij})^2}.
\]
\end{proposition}

\begin{corollary}[Same-GroupSort pairwise bound]
\label{cor:same}
If $\Pi^{(\ell)}_i = \Pi^{(\ell)}_j$, then
\[
  \left|d^{(\ell+1)}_{ij} - d^{(\ell)}_{ij}\right|
  \;\le\;
  2\rho\sqrt{1 - (\gamma^{(\ell)}_{ij})^2}.
\]
Summing over layers:
\[
  d^{(L)}_{ij} \;\ge\; d^{(0)}_{ij}
  - 2\rho\sum_{\ell=0}^{L-1}\sqrt{1-(\gamma^{(\ell)}_{ij})^2}.
\]
\end{corollary}

\begin{remark}
If $\sum_{\ell=0}^{\infty}\sqrt{1-(\gamma^{(\ell)}_{ij})^2}
< d^{(0)}_{ij}/(2\rho)$, then nodes $i$ and $j$ can never collapse to the same
hidden vector, regardless of depth.  Conversely, for collapse not to be ruled out
by this bound, the cumulative angular mismatch must eventually reach at least this
threshold.  In particular, if two nodes receive collinear messages
($\gamma^{(\ell)}_{ij}=1$) at every layer, their distance is preserved exactly.
\end{remark}

\subsection{Role of the Adjacency Spectrum}

The adjacency operator $P$ still shapes the messages $\mathbf{m}^{(\ell)}_i$ in
\eqref{eq:aggregate}---its eigenvalues influence the direction field
$\{\mathbf{u}^{(\ell)}_i\}$.  What Proposition~\ref{prop:scale} proves is that
\emph{message magnitude} is discarded before the state update.  The standard
oversmoothing mechanism---in which eigenvalues of $S$ appear as repeated
multiplicative factors on the hidden state---is absent.  Oversmoothing in
HouseGNN, if it occurs, must occur through accumulated node-wise operator
mismatches that gradually rotate or reflect representations toward one another.
If all nodes experience the same local orthogonal operator, pairwise distances are
preserved rather than smoothed.

\section{Proofs}
\label{sec:proofs}

\subsection{Proof of Proposition~\ref{prop:ortho}}

Let $\mathbf{u}\in\R^d$ satisfy $\norm{\mathbf{u}}_2=1$, and define
\begin{equation}
  R = \Id - 2\mathbf{u}\mathbf{u}^{\top}.
\end{equation}

\textit{Symmetry.}
Taking the transpose gives
\begin{equation}
  R^{\top}
  =
  \left(\Id - 2\mathbf{u}\mathbf{u}^{\top}\right)^{\top}
  =
  \Id - 2\mathbf{u}\mathbf{u}^{\top}
  =
  R .
\end{equation}
Hence, $R$ is symmetric.

\textit{Orthogonality.}
We compute
\begin{align}
  R^2
  &=
  \left(\Id - 2\mathbf{u}\mathbf{u}^{\top}\right)^2  \nonumber\\
  &=
  \Id
  -
  4\mathbf{u}\mathbf{u}^{\top}
  +
  4\mathbf{u}
  \left(\mathbf{u}^{\top}\mathbf{u}\right)
  \mathbf{u}^{\top}.
\end{align}
Since $\mathbf{u}^{\top}\mathbf{u}=1$, this reduces to
\begin{equation}
  R^2 = \Id .
\end{equation}
Because $R$ is symmetric, we have
\begin{equation}
  R^{\top}R = R^2 = \Id .
\end{equation}
Therefore, $R$ is orthogonal.

\textit{Eigenvalues.}
First,
\begin{equation}
  R\mathbf{u}
  =
  \mathbf{u}
  -
  2\mathbf{u}
  \left(\mathbf{u}^{\top}\mathbf{u}\right)
  =
  -\mathbf{u}.
\end{equation}
Thus, $\mathbf{u}$ is an eigenvector with eigenvalue $-1$.

Now let $\mathbf{w}\perp\mathbf{u}$. Then
$\mathbf{u}^{\top}\mathbf{w}=0$, and hence
\begin{equation}
  R\mathbf{w}
  =
  \mathbf{w}
  -
  2\mathbf{u}
  \left(\mathbf{u}^{\top}\mathbf{w}\right)
  =
  \mathbf{w}.
\end{equation}
Therefore, every vector that satisfies $\mathbf{u}^{\top}\mathbf{w}=0$ is an eigenvector with
eigenvalue $+1$. This proves the proposition.
\hfill$\square$

\subsection{Proof of Proposition~\ref{prop:norm}}

Let
\begin{equation}
  \mathbf{z}^{(\ell)}_i
  =
  R_i^{(\ell)}\mathbf{h}^{(\ell)}_i .
\end{equation}
By Proposition~\ref{prop:ortho}, $R_i^{(\ell)}$ is orthogonal. Therefore,
\begin{equation}
  \norm{\mathbf{z}^{(\ell)}_i}_2
  =
  \norm{\mathbf{h}^{(\ell)}_i}_2 .
\end{equation}

GroupSort only permutes coordinates within each group. Since coordinate
permutations preserve the Euclidean norm,
\begin{equation}
  \norm{
  \mathrm{GroupSort}
  \left(\mathbf{z}^{(\ell)}_i\right)
  }_2
  =
  \norm{\mathbf{z}^{(\ell)}_i}_2 .
\end{equation}
Using the update rule
\[
  \mathbf{h}^{(\ell+1)}_i
  =
  \mathrm{GroupSort}
  \left(\mathbf{z}^{(\ell)}_i\right),
\]
we obtain
\begin{equation}
  \norm{\mathbf{h}^{(\ell+1)}_i}_2
  =
  \norm{\mathbf{h}^{(\ell)}_i}_2 .
\end{equation}
Applying this equality recursively gives
\begin{equation}
  \norm{\mathbf{h}^{(\ell)}_i}_2
  =
  \norm{\mathbf{h}^{(0)}_i}_2,
  \qquad
  \ell\ge 0 .
\end{equation}
This proves the proposition.
\hfill$\square$

\subsection{Proof of Proposition~\ref{prop:scale}}

Let $\mathbf{v}\neq\mathbf{0}$ and let $c\neq0$. Then
\begin{equation}
  \frac{c\mathbf{v}}{\norm{c\mathbf{v}}_2}
  =
  \frac{c}{|c|}
  \frac{\mathbf{v}}{\norm{\mathbf{v}}_2}
  =
  \mathrm{sgn}(c)
  \frac{\mathbf{v}}{\norm{\mathbf{v}}_2}.
\end{equation}
The scalar factor disappears in the outer product, since
$\mathrm{sgn}(c)^2=1$. Hence,
\begin{align}
  R(c\mathbf{v})
  &=
  \Id
  -
  2
  \left(
  \frac{c\mathbf{v}}{\norm{c\mathbf{v}}_2}
  \right)
  \left(
  \frac{c\mathbf{v}}{\norm{c\mathbf{v}}_2}
  \right)^{\top} \nonumber\\
  &=
  \Id
  -
  2
  \frac{\mathbf{v}}{\norm{\mathbf{v}}_2}
  \frac{\mathbf{v}^{\top}}{\norm{\mathbf{v}}_2}
  =
  R(\mathbf{v}).
\end{align}
This proves scale and sign invariance.
\hfill$\square$

\subsection{Proof of Proposition~\ref{prop:collapse}}

Suppose
\begin{equation}
  \mathbf{h}^{(L)}_i
  =
  \mathbf{h}^{(L)}_j .
\end{equation}
Then their Euclidean norms are equal:
\begin{equation}
  \norm{\mathbf{h}^{(L)}_i}_2
  =
  \norm{\mathbf{h}^{(L)}_j}_2 .
\end{equation}
By Proposition~\ref{prop:norm},
\begin{equation}
  \norm{\mathbf{h}^{(L)}_k}_2
  =
  \norm{\mathbf{h}^{(0)}_k}_2,
  \qquad
  k\in\{i,j\}.
\end{equation}
Therefore,
\begin{equation}
  \norm{\mathbf{h}^{(0)}_i}_2
  =
  \norm{\mathbf{h}^{(0)}_j}_2 .
\end{equation}
This proves the proposition.
\hfill$\square$

\subsection{Proof of Theorem~\ref{thm:distance}}

Since $O_i^{(\ell)}$ and $O_j^{(\ell)}$ are orthogonal, we have
\begin{align}
  \Delta_{ij}^{(\ell+1)}
  &=
  O_i^{(\ell)}\mathbf{h}^{(\ell)}_i
  -
  O_j^{(\ell)}\mathbf{h}^{(\ell)}_j
  \nonumber\\
  &=
  O_i^{(\ell)}
  \left(
  \mathbf{h}^{(\ell)}_i
  -
  \mathbf{h}^{(\ell)}_j
  \right)
  +
  \left(
  O_i^{(\ell)}
  -
  O_j^{(\ell)}
  \right)
  \mathbf{h}^{(\ell)}_j .
\end{align}
Taking norms and applying the triangle inequality gives
\begin{align}
  d_{ij}^{(\ell+1)}
  &\le
  \norm{
  O_i^{(\ell)}
  \Delta_{ij}^{(\ell)}
  }_2
  +
  \norm{
  \left(
  O_i^{(\ell)}
  -
  O_j^{(\ell)}
  \right)
  \mathbf{h}^{(\ell)}_j
  }_2
  \nonumber\\
  &\le
  \norm{\Delta_{ij}^{(\ell)}}_2
  +
  \norm{
  O_i^{(\ell)}
  -
  O_j^{(\ell)}
  }_2
  \norm{\mathbf{h}^{(\ell)}_j}_2 .
\end{align}
Because $O_i^{(\ell)}$ is orthogonal,
\[
  \norm{
  O_i^{(\ell)}
  \Delta_{ij}^{(\ell)}
  }_2
  =
  \norm{\Delta_{ij}^{(\ell)}}_2
  =
  d_{ij}^{(\ell)} .
\]
Moreover, by Proposition~\ref{prop:norm},
\begin{equation}
  \norm{\mathbf{h}^{(\ell)}_j}_2
  =
  \norm{\mathbf{h}^{(0)}_j}_2
  \le
  \rho .
\end{equation}
Therefore,
\begin{equation}
  d_{ij}^{(\ell+1)}
  \le
  d_{ij}^{(\ell)}
  +
  \rho
  \norm{
  O_i^{(\ell)}
  -
  O_j^{(\ell)}
  }_2 .
\end{equation}
Equivalently,
\begin{equation}
  d_{ij}^{(\ell+1)}
  -
  d_{ij}^{(\ell)}
  \le
  \rho
  \norm{
  O_i^{(\ell)}
  -
  O_j^{(\ell)}
  }_2 .
\end{equation}

For the lower bound, apply the reverse triangle inequality to the same
decomposition:
\begin{equation}
\begin{aligned}
\norm{\Delta_{ij}^{(\ell+1)}}_2
&\ge
\norm{O_i^{(\ell)}\Delta_{ij}^{(\ell)}}_2
-
\norm{
\left(
O_i^{(\ell)}-O_j^{(\ell)}
\right)
\mathbf{h}_j^{(\ell)}
}_2 .
\end{aligned}
\end{equation}
Using the same estimates,
\begin{equation}
  d_{ij}^{(\ell+1)}
  \ge
  d_{ij}^{(\ell)}
  -
  \rho
  \norm{
  O_i^{(\ell)}-O_j^{(\ell)}
  }_2 .
\end{equation}
Combining the upper and lower bounds yields
\begin{equation}
  \left|
  d_{ij}^{(\ell+1)}
  -
  d_{ij}^{(\ell)}
  \right|
  \le
  \rho
  \norm{
  O_i^{(\ell)}-O_j^{(\ell)}
  }_2 .
\end{equation}
This proves the theorem.
\hfill$\square$

\subsection{Proof of Proposition~\ref{prop:mismatch}}

Let $\mathbf{u},\mathbf{w}\in\R^d$ be unit vectors and define
\begin{equation}
  c=\mathbf{u}^{\top}\mathbf{w}.
\end{equation}
If $c=1$, then $\mathbf{u}=\mathbf{w}$. If $c=-1$, then
$\mathbf{w}=-\mathbf{u}$. In both cases,
\[
  \mathbf{w}\mathbf{w}^{\top}
  =
  \mathbf{u}\mathbf{u}^{\top},
\]
so $R(\mathbf{u})=R(\mathbf{w})$ and the result is immediate.

Now assume $|c|<1$. We have
\begin{equation}
  R(\mathbf{u})-R(\mathbf{w})
  =
  2
  \left(
  \mathbf{w}\mathbf{w}^{\top}
  -
  \mathbf{u}\mathbf{u}^{\top}
  \right).
\end{equation}
Define
\begin{equation}
  \mathbf{s}
  =
  \frac{
  \mathbf{w}
  -
  c\mathbf{u}
  }
  {
  \sqrt{1-c^2}
  } .
\end{equation}
Then $\{\mathbf{u},\mathbf{s}\}$ is an orthonormal set, and
\begin{equation}
  \mathbf{w}
  =
  c\mathbf{u}
  +
  \sqrt{1-c^2}\,\mathbf{s}.
\end{equation}
In the basis $\{\mathbf{u},\mathbf{s}\}$, the matrix
$\mathbf{w}\mathbf{w}^{\top}-\mathbf{u}\mathbf{u}^{\top}$ has representation
\begin{equation}
  \begin{pmatrix}
    -(1-c^2) & c\sqrt{1-c^2} \\
    c\sqrt{1-c^2} & 1-c^2
  \end{pmatrix}.
\end{equation}
This symmetric matrix has trace zero and determinant $-(1-c^2)$. Hence, its
eigenvalues are
\begin{equation}
  \pm\sqrt{1-c^2}.
\end{equation}
Therefore,
\begin{equation}
  \norm{
  \mathbf{w}\mathbf{w}^{\top}
  -
  \mathbf{u}\mathbf{u}^{\top}
  }_2
  =
  \sqrt{1-c^2}.
\end{equation}
It follows that
\begin{equation}
  \norm{
  R(\mathbf{u})-R(\mathbf{w})
  }_2
  =
  2\sqrt{1-c^2}.
\end{equation}
Since $\gamma_{ij}^{(\ell)}=|c|$, we have
$c^2=(\gamma_{ij}^{(\ell)})^2$. Thus,
\begin{equation}
  \norm{
  R_i^{(\ell)}-R_j^{(\ell)}
  }_2
  =
  2
  \sqrt{
  1-(\gamma_{ij}^{(\ell)})^2
  } .
\end{equation}
This proves the proposition.
\hfill$\square$

\section{Experimental Evaluation}
\label{sec:experiments}

\subsection{Experimental Protocol}

\textbf{Datasets.}  We evaluate on six standard node-classification benchmarks.
The citation graphs Cora~\cite{mccallum2000automating},
CiteSeer~\cite{giles1998citeseer}, and PubMed~\cite{namata2012query} are
homophilic (edge homophily $\ge 0.74$).  Texas, Wisconsin, and Cornell from the
WebKB collection are heterophilic (edge homophily $\le 0.30$).  Dataset
statistics are provided in Table~\ref{tab:datasets}.

\begin{table}[h]
\centering
\caption{Dataset statistics.  $H(\mathcal{G})$ denotes edge homophily.}
\label{tab:datasets}
\begin{tabular}{lrrrc}
\toprule
Dataset & Nodes & Edges & Classes & $H(\mathcal{G})$ \\
\midrule
Cora      & 2{,}708 & 5{,}429  & 7  & 0.81 \\
CiteSeer  & 3{,}327 & 4{,}732  & 6  & 0.74 \\
PubMed    & 19{,}717 & 44{,}338 & 3  & 0.80 \\
Texas     & 183   & 295    & 5  & 0.21 \\
Wisconsin & 251   & 466    & 5  & 0.11 \\
Cornell   & 183   & 280    & 5  & 0.30 \\
\bottomrule
\end{tabular}
\end{table}

\textbf{Baselines.}  We compare against (i) a standard GCN~\cite{kipf2017semi},
(ii) GAT~\cite{velickovic2018gat}, (iii) BatchNorm-GCN~\cite{ioffe2015batch},
(iv) PairNorm-GCN~\cite{zhao2020pairnorm}, and (v) Residual-GCN.  Baseline
results for GCN, GAT, BatchNorm, PairNorm, and Residual on Cora, CiteSeer, and
PubMed are taken from Wang~et~al.~\cite{wang2025signed}.

\textbf{HouseGNN setup.}  Hidden dimension: 64; GroupSort group size: 2;
aggregation: mean; optimiser: Adam with learning rate $5\times10^{-4}$ and weight
decay $10^{-3}$; training: up to 400 epochs with gradient clipping. All HouseGNN results are reported as the mean and standard deviation over 5 random seeds.

\textbf{Depth sweep.}  The main benchmark uses $L\in\{2,8,16,32,64\}$ layers.
In addition, we include a Cora-only geometric diagnostic across the same depth
range and an extreme-depth study at $L=128$ comparing GroupSort and ReLU under
the same orthogonal Householder backbone.

\subsection{Core Benchmark: Cora, CiteSeer, and Texas}

\begin{figure*}[t]
  \centering
  \begin{subfigure}[b]{0.32\textwidth}
    \safeincludegraphics[width=\linewidth]{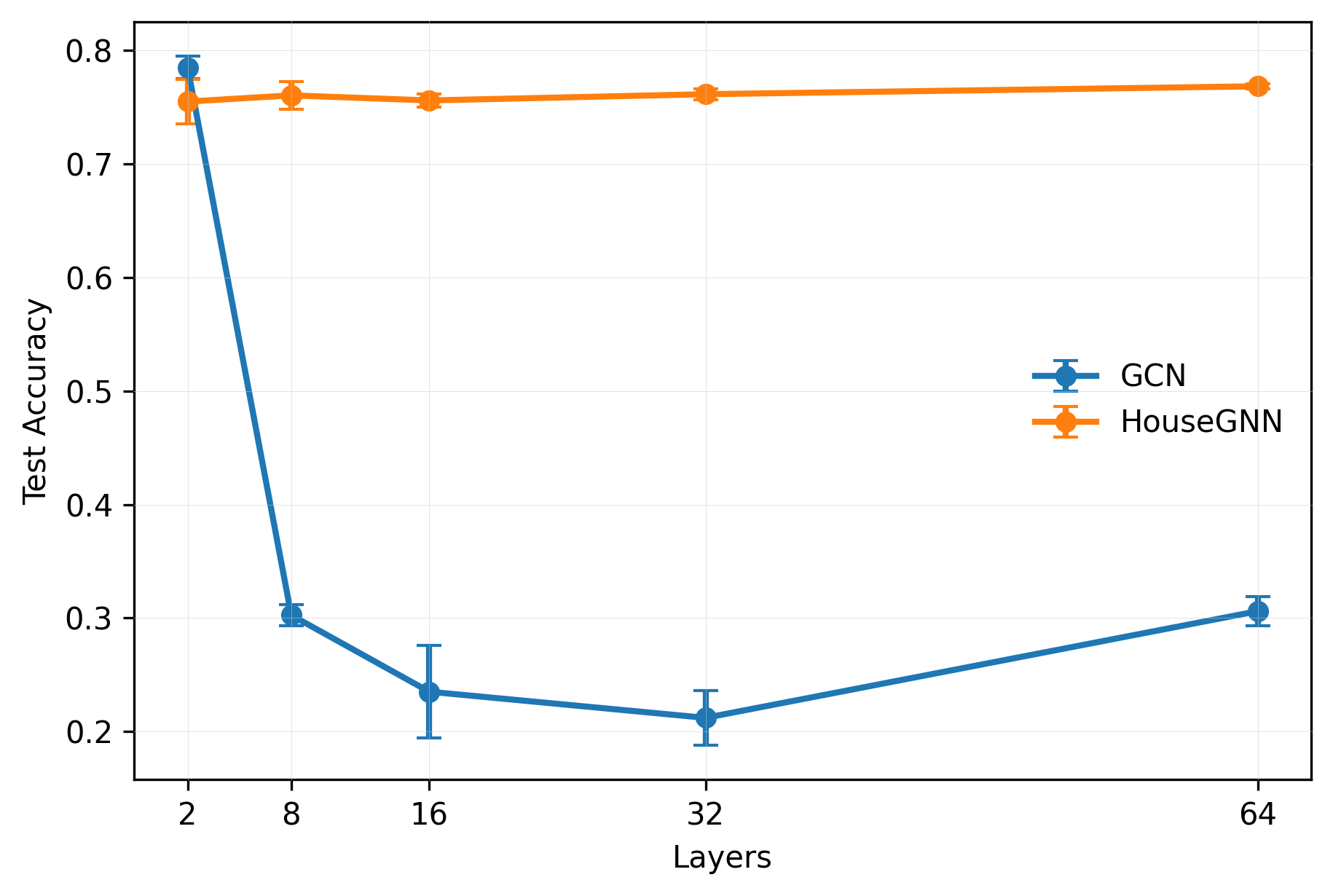}
    \caption{Cora}
    \label{fig:cora}
  \end{subfigure}
  \hfill
  \begin{subfigure}[b]{0.32\textwidth}
    \safeincludegraphics[width=\linewidth]{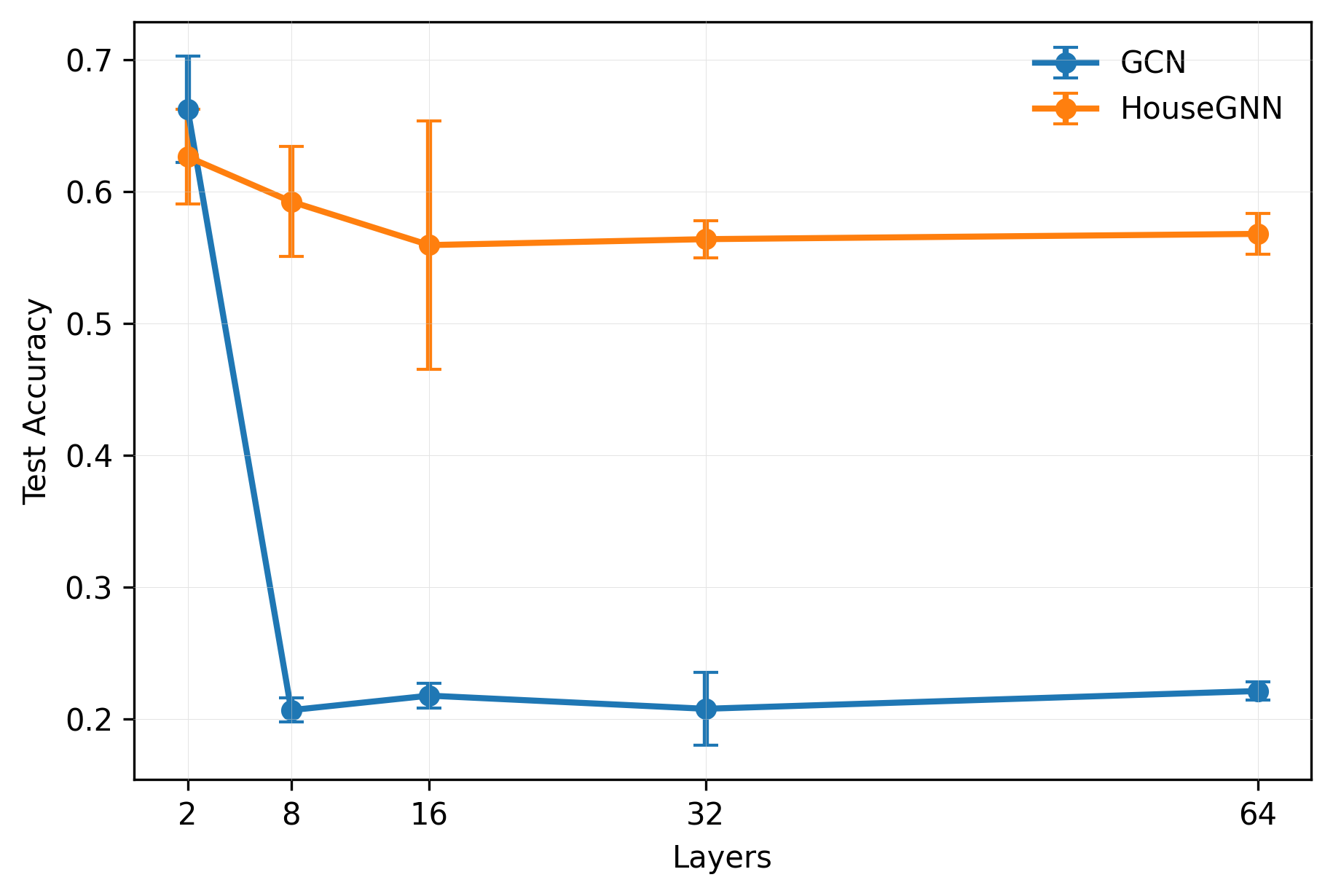}
    \caption{CiteSeer}
    \label{fig:citeseer}
  \end{subfigure}
  \hfill
  \begin{subfigure}[b]{0.32\textwidth}
    \safeincludegraphics[width=\linewidth]{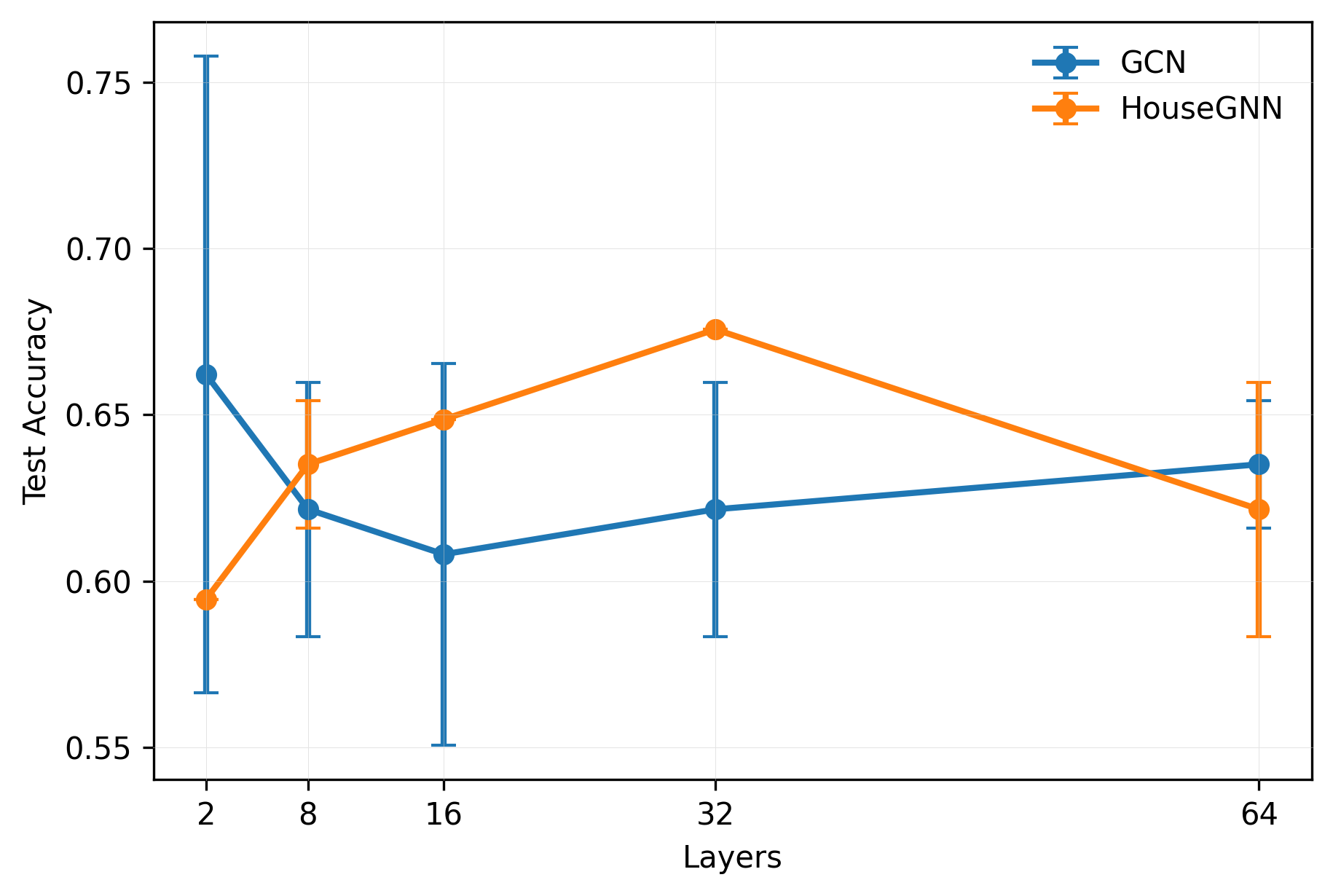}
    \caption{Texas}
    \label{fig:texas}
  \end{subfigure}
  \caption{Test accuracy vs.\ depth for GCN (blue) and HouseGNN (orange) on
  Cora, CiteSeer, and Texas.}
  \label{fig:core}
\end{figure*}

Figure~\ref{fig:core} shows the test accuracy curves as the depth varies.
This shows: GCN collapses
beyond two layers, while HouseGNN maintains stable accuracy across the full
depth sweeps.

Table~\ref{tab:core} contains numerical results.  \textbf{Cora}: GCN accuracy
drops from $78.5\%$ at 2 layers to around $21$--$30\%$ at $8$--$64$ layers.
HouseGNN stays in the
$75.5$--$76.8\%$ range throughout. \textbf{CiteSeer}: GCN collapses from
$66.2\%$ at 2 layers to below $22\%$ at $8$--$64$ layers; HouseGNN varies between
$55.9\%$ to $62.7\%$ throughout.  \textbf{Texas}: Both models remain comparatively stable, consistent with the heterophilic structure of the dataset.

\begin{table*}[t]
\centering
\caption{Core benchmark: Test accuracy vs \ layers on Cora, CiteSeer, and
Texas.}
\label{tab:core}
\setlength{\tabcolsep}{4.5pt}
\begin{tabular}{ll ccccc}
\toprule
\multirow{2}{*}{Dataset} & \multirow{2}{*}{Model}
  & \multicolumn{5}{c}{Number of Layers} \\
\cmidrule(lr){3-7}
  & & $L=2$ & $L=8$ & $L=16$ & $L=32$ & $L=64$ \\
\midrule
\multirow{2}{*}{Cora}
  & GCN~\cite{kipf2017semi}
    & $78.5\pm1.0$ & $30.2\pm0.9$ & $23.5\pm4.1$ & $21.2\pm2.4$ & $30.6\pm1.3$ \\
  & \textbf{HouseGNN (Ours)}
    & $75.5\pm2.0$ & $76.0\pm1.2$ & $75.6\pm0.6$
    & $76.2\pm0.5$ & $76.8\pm0.2$ \\
\midrule
\multirow{2}{*}{CiteSeer}
  & GCN~\cite{kipf2017semi}
    & $66.2\pm4.0$ & $20.6\pm0.9$ & $21.7\pm0.9$ & $20.7\pm2.8$ & $22.1\pm0.7$ \\
  & \textbf{HouseGNN (Ours)}
    & $62.7\pm3.6$ & $59.3\pm4.2$ & $55.9\pm9.4$
    & $56.4\pm1.4$ & $56.8\pm1.6$ \\
\midrule
\multirow{2}{*}{Texas}
  & GCN~\cite{kipf2017semi}
    & $66.2\pm9.6$ & $62.2\pm3.8$ & $60.8\pm5.7$ & $62.2\pm3.8$ & $63.5\pm1.9$ \\
  & \textbf{HouseGNN (Ours)}
    & $59.5\pm0.0$ & $63.5\pm1.9$ & $64.9\pm0.0$ & $67.6\pm0.0$ & $62.2\pm3.8$ \\
\bottomrule
\end{tabular}
\end{table*}
\subsection{Comparison with Multiple Baselines}

Table~\ref{tab:full} provides a broad comparison on Cora, CiteSeer, and PubMed,
including GCN, GAT, BatchNorm, PairNorm, and Residual.  Baseline results are
taken from~\cite{wang2025signed}.

\textbf{GAT.}  On Cora, GAT accuracy falls from
$81.5\%$ at $L=2$ to $28.4\%$ at $L=64$.
On CiteSeer, GAT drops from $69.9\%$ to $26.0\%$.

\textbf{Partial mitigation by BatchNorm and PairNorm.}  These normalisations slow
the collapse but do not eliminate it.  At $L=64$ on Cora, BatchNorm achieves only
$35.3\%$ and PairNorm $44.0\%$.  On PubMed, both methods are more effective
($69.3\%$ and $71.2\%$ at $L=64$ respectively), but still degrade notably from
their peak values.

\textbf{Residual connections} provide limited depth-robustness: on Cora, accuracy
drops from $80.7\%$ at $L=2$ to $27.9\%$ at $L=64$; on CiteSeer it collapses
from $67.5\%$ at $L=2$ to $19.7\%$ at $L=64$.

\textbf{HouseGNN} shows the most stable depth trend among the reported methods.  On Cora, it
maintains $75.5$--$76.8\%$ across all tested depths.  On CiteSeer, it ranges
between $55.9\%$ and $62.7\%$.  On PubMed, performance reaches $77.0\%$ at $L=64$. 

\begin{table*}[t]
\centering
\caption{Test accuracy (\%) vs \ depth on Cora, CiteSeer, and
PubMed.  Baseline results ( GAT, BatchNorm, PairNorm, Residual) are from
Wang~et~al.~\cite{wang2025signed}. }
\label{tab:full}
\setlength{\tabcolsep}{3.5pt}
\begin{tabular}{ll ccccc}
\toprule
Dataset & Model
  & $L=2$ & $L=8$ & $L=16$ & $L=32$ & $L=64$ \\
\midrule
\multirow{6}{*}{Cora}
  & GCN~\cite{kipf2017semi}
    & $78.5\pm1.0$ & $30.2\pm0.9$ & $23.5\pm4.1$ & $21.2\pm2.4$ & $30.6\pm1.3$ \\
  & GAT~\cite{velickovic2018gat}
    & $81.5\pm0.5$ & $58.6\pm2.0$ & $25.2\pm5.7$ & $31.9\pm0.2$ & $28.4\pm0.0$ \\
  & BatchNorm~\cite{ioffe2015batch}
    & $78.1\pm0.0$ & $73.6\pm0.6$ & $70.8\pm0.0$ & $53.9\pm2.2$ & $35.3\pm3.4$ \\
  & PairNorm~\cite{zhao2020pairnorm}
    & $79.0\pm0.0$ & $73.2\pm0.0$ & $63.0\pm0.0$ & $48.1\pm0.9$ & $44.0\pm3.5$ \\
  & Residual~\cite{he2016deep}
    & $80.7\pm0.1$ & $79.3\pm0.2$ & $40.9\pm0.0$ & $31.0\pm0.0$ & $27.9\pm6.1$ \\
  & \textbf{HouseGNN (Ours)}
    & $75.5\pm2.0$
    & $76.0\pm1.2$
    & $75.6\pm0.6$
    & $76.2\pm0.5$
    & $76.8\pm0.2$ \\
\midrule
\multirow{6}{*}{CiteSeer}
  & GCN~\cite{kipf2017semi}
    & $66.2\pm4.0$ & $20.6\pm0.9$ & $21.7\pm0.9$ & $20.7\pm2.8$ & $22.1\pm0.7$ \\
  & GAT~\cite{velickovic2018gat}
    & $69.9\pm0.9$ & $44.7\pm3.1$ & $23.5\pm1.4$ & $24.4\pm0.4$ & $26.0\pm2.2$ \\
  & BatchNorm~\cite{ioffe2015batch}
    & $63.4\pm0.9$ & $61.4\pm0.0$ & $50.6\pm1.2$ & $41.4\pm0.0$ & $35.0\pm1.1$ \\
  & PairNorm~\cite{zhao2020pairnorm}
    & $63.6\pm0.6$ & $62.0\pm1.2$ & $50.1\pm0.0$ & $37.2\pm1.9$ & $36.1\pm0.1$ \\
  & Residual~\cite{he2016deep}
    & $67.5\pm0.5$ & $67.3\pm0.0$ & $33.2\pm0.0$ & $19.7\pm0.0$ & $19.7\pm0.0$ \\
  & \textbf{HouseGNN (Ours)}
    & $62.7\pm3.6$
    & $59.3\pm4.2$
    & $55.9\pm9.4$
    & $56.4\pm1.4$
    & $56.8\pm1.6$ \\
\midrule
\multirow{5}{*}{PubMed}
  & GCN~\cite{kipf2017semi}
    & $76.4\pm0.3$ & $69.6\pm5.9$ & $39.9\pm0.0$ & $39.9\pm0.0$ & $39.9\pm0.0$ \\
  & BatchNorm~\cite{ioffe2015batch}
    & $75.5\pm0.1$ & $77.1\pm0.0$ & $76.9\pm0.0$ & $75.4\pm0.0$ & $69.3\pm1.0$ \\
  & PairNorm~\cite{zhao2020pairnorm}
    & $75.7\pm0.1$ & $78.0\pm0.0$ & $77.2\pm0.4$ & $75.5\pm2.0$ & $71.2\pm3.7$ \\
  & Residual~\cite{he2016deep}
    & $76.4\pm0.3$ & $77.4\pm0.0$ & $63.1\pm3.1$ & $39.9\pm0.0$ & $39.9\pm0.0$ \\
  & \textbf{HouseGNN (Ours)}
    & $71.1\pm4.3$
    & $72.6\pm2.2$
    & $70.7\pm2.4$
    & $73.8\pm2.5$
    & $77.0\pm1.6$ \\
\bottomrule
\end{tabular}
\end{table*}

\subsection{Extended Experiments: Wisconsin and Cornell}

Table~\ref{tab:extended} reports HouseGNN on Wisconsin and Cornell.

\begin{table}[!t]
\centering
\caption{Result on Wisconsin and Cornell Dataset for HouseGNN.}
\label{tab:extended}
\begin{tabular}{llc}
\toprule
\text{Dataset} & \text{L} & \text{Test Accuracy} \\
\midrule
\multirow{5}{*}{Wisconsin}
  & 2  & $61.4\pm15.8$ \\
  & 8  & $61.4\pm7.9$ \\
  & 16 & $62.1\pm7.9$ \\
  & 32 & $62.1\pm13.1$ \\
  & 64 & $58.2\pm6.0$ \\
\midrule
\multirow{5}{*}{Cornell}
  & 2  & $50.5\pm6.2$ \\
  & 8  & $55.9\pm5.6$ \\
  & 16 & $53.2\pm3.1$ \\
  & 32 & $55.9\pm8.3$ \\
  & 64 & $51.4\pm8.1$ \\
\bottomrule
\end{tabular}
\end{table}
\textbf{Wisconsin.}  The model is stable across depth, attaining its best
test accuracy at 16 and 32 layers ($62.1\%$).
\textbf{Cornell.}  Best performance appears at 8 and 32 layers ($55.9\%$).

\subsection{Removing the Orthogonal Weights}

To isolate the contribution of orthogonal weights, we evaluate HouseGNN on Cora at 64 and 80 layers where the
matrix $W^{(\ell)}$ in \eqref{eq:project} is unconstrained (not required to have
orthogonal weights).  The Householder reflection and GroupSort remain intact.
Table~\ref{tab:ablation} reports the results for this.

\begin{table}[h]
\centering
\caption{Result on Cora after removing the orthogonal weights constraint on HouseGNN.}
\label{tab:ablation}
\setlength{\tabcolsep}{5pt}
\begin{tabular}{lcrr}
\toprule
\text{Variant} & $L$ & \text{Test Acc.} &  \\
\midrule
Orthogonal (reference) & 64 & $76.8\pm0.2$ \\
No orthogonal constraint & 64 & $76.4\pm1.8$  \\
No orthogonal constraint & 80 & $76.7\pm1.7$  \\
\bottomrule
\end{tabular}
\end{table}

These results suggest that the model does not collapse at large depth when the
orthogonality constraint is removed, suggesting that the Householder plus GroupSort activation itself carries meaningful learning capacity.

\subsection{Dirichlet Energy and effective rank on Cora}

\begin{figure*}[t]
  \centering
  \begin{subfigure}[b]{0.48\textwidth}
    \safeincludegraphics[width=\linewidth]{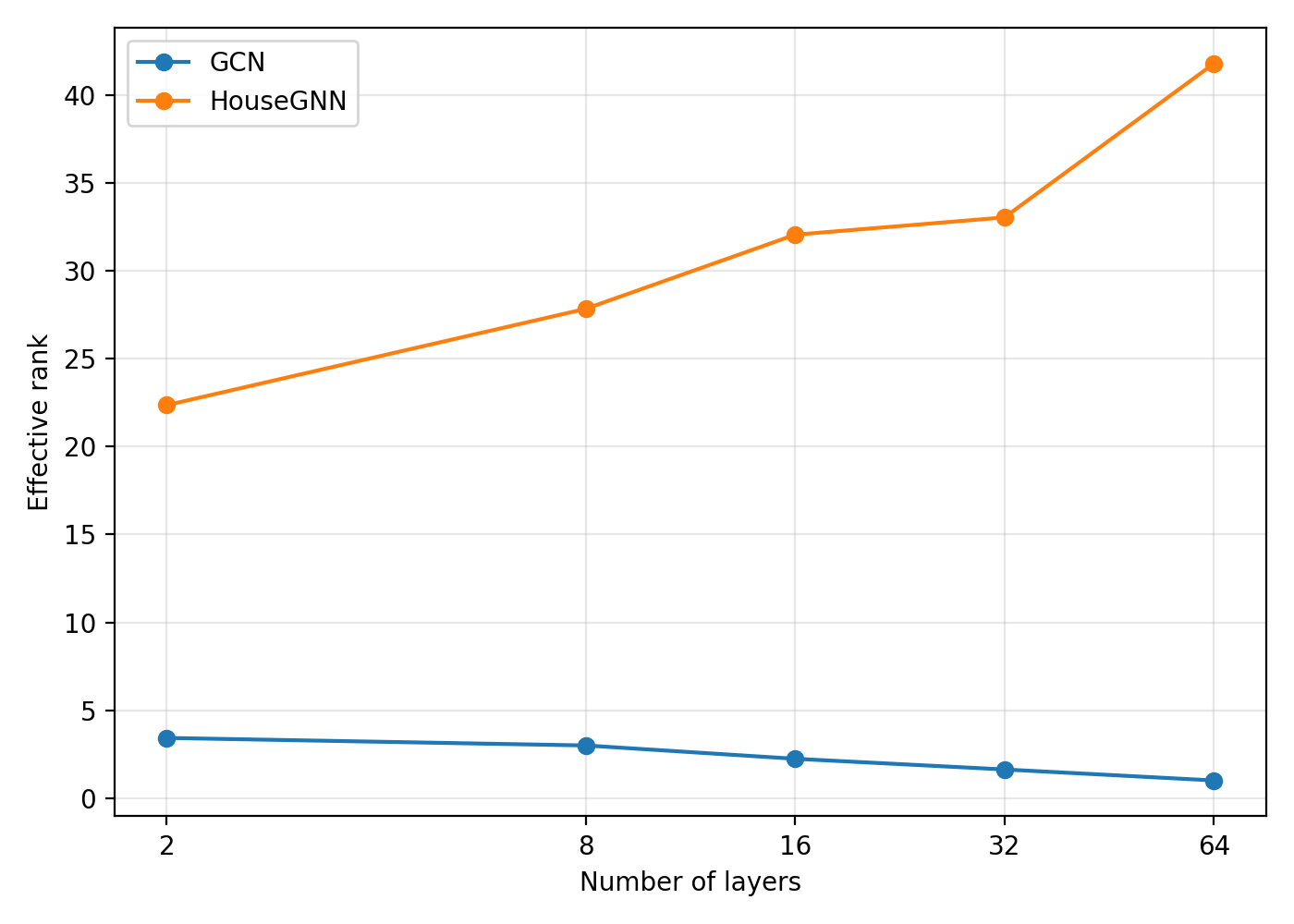}

    \label{fig:rank_depth}
  \end{subfigure}
  \hfill
  \begin{subfigure}[b]{0.48\textwidth}
    \safeincludegraphics[width=\linewidth]{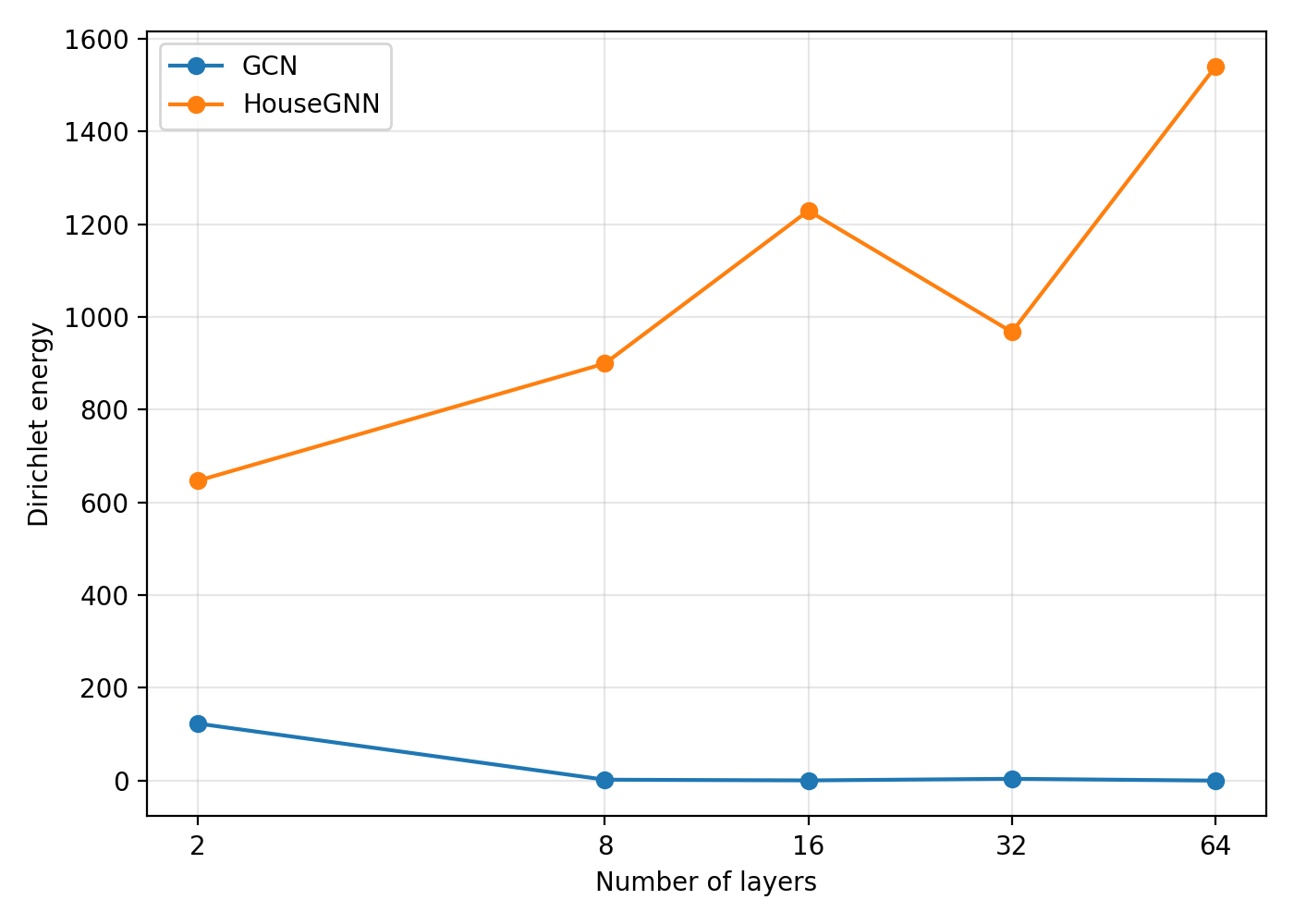}

    \label{fig:dirichlet_depth}
  \end{subfigure}
  \caption{Effective rank and Dirichlet energy vs . depth on Cora}
  \label{fig:geometry}
\end{figure*}

Figure~\ref{fig:geometry} shows that GCN rapidly loses Dirichlet energy as
depth increases, and its effective rank also remains very low. In contrast,
HouseGNN maintains a much larger effective rank throughout the sweep and does
not show the same collapse of Dirichlet energy. Dirichlet energy is used as a
metric for oversmoothing; therefore, for GCN, as the number of layers increases,
both accuracy and Dirichlet energy decrease.

\subsection{Result at 128-Layers}
Table~\ref{tab:128relu} reports the Cora results
on 128 layers with two activation functions.

\begin{table}[!t]
\centering
\caption{Result on Cora at 128-layers on HouseGNN.}
\label{tab:128relu}
\setlength{\tabcolsep}{6pt}
\begin{tabular}{lccc}
\toprule
Activation & Test acc. (\%) & Effective rank & Dirichlet energy \\
\midrule
ReLU      & $71.9\pm1.6$ & $25.148\pm2.450$ & $658.764\pm43.253$ \\
GroupSort & $6.4\pm0.1$  & $54.014\pm0.413$ & $3506.059\pm152.905$ \\
\bottomrule
\end{tabular}
\end{table}

As suggested by the results in Table~\ref{tab:128relu} the  contrast is important. At 128 layers, HouseGNN with GroupSort activation  falls
to very low test accuracy, but its effective rank remains high and its
Dirichlet energy also remains large (ReLU: $658.764\pm43.253$; GroupSort:
$3506.059\pm152.905$). This indicates that the 128-layer failure is not well
described as classical GCN-style oversmoothing, where one would expect
representational collapse and a large decrease in Dirichlet energy. One possible interpretation is that the orthogonal Householder+GroupSort dynamics
can enter a non-collapsed but poorly class-discriminative regime at extreme
depth.

Another possible explanation is that, at 128 layers, repeated reflections combined
with GroupSort may make the representations overly dispersed or poorly aligned
for classification. Since GroupSort permutes the elements of each vector while
preserving its norm, the representations may remain non-collapsed but become
poorly organized in direction. This interpretation is supported by the high
effective rank, which suggests that the representations occupy many
directions in the feature space.

This is also consistent with the large Dirichlet energy, which suggests that
node representations do not collapse but instead remain highly separated in the
representation space. In this regime, more different activation may help
organize the representations into a more class-discriminative structure. The 128-layers result of the ReLU variant supports this interpretation,
although it should be viewed as suggestive rather than conclusive.

\subsection{Summary of Empirical Findings}

Conclusions from the experiments:

\begin{enumerate}
  \item \textbf{Depth stability.} HouseGNN maintains
        relatively stable accuracy as depth increases, whereas GCN shows a
        substantial drop at larger depths.

  \item \textbf{Behaviour across datasets.}  The results on PubMed, Wisconsin,
        and Cornell suggest that the proposed update remains stable across both
        homophilic and heterophilic graphs.

  \item \textbf{Effect of the orthogonality constraint.}  The Cora ablation shows
        that removing the orthogonal constraint does not lead to collapse,
        suggesting that the reflection-based update itself plays an important
        role in depth stability.

\end{enumerate}

\section{Conclusion and Future Work}
\label{sec:conclusion}

This work studied Householder Graph Neural Network (HouseGNN) as a possible way
to mitigate oversmoothing in deep GNNs. The main idea is to
use graph aggregation to estimate a direction, rather than directly using the aggregated
message as the new hidden state. The node state
is updated by a Householder reflection followed
by GroupSort, so each internal layer is piecewise
orthogonal and preserves the Euclidean norm. The theoretical analysis shows HouseGNN layer preserves
node-wise norms at every depth and pairwise distances can change only through
mismatch between local orthogonal operators.  These results do not prove that all
forms of oversmoothing are impossible, but they show that the standard
diffusion-based mechanism of GCN oversmoothing is removed.

The experiments support the theory.  HouseGNN remains stable across many layers
on Cora and CiteSeer, while GCN collapses after only a few layers.  On Texas,
Wisconsin, and Cornell, HouseGNN shows mixed behaviour, which suggests that
heterophilic graphs need additional modelling choices.  On PubMed, the model
shows promising behaviour at larger depth. The additional diagnostics
on Cora further indicate that HouseGNN does not follow the usual low-rank,
low-energy oversmoothing pattern of GCN across the depth sweep.

At the same time, the 128-layer result shows the model can fail but failure can be migigated by changing the activation function. The GroupSort version can become non-discriminative without
showing collapse in effective rank and dirichlet energy, whereas the ReLU version
remains stable.

Future work should investigate whether the proposed method can be extended
beyond 128 layers. During our experiments, we tested a 150-layer model with
ReLU activations and observed accuracy comparable to that of the 128-layer
ReLU model, suggesting that deeper architectures may be feasible. Another
promising direction is to study the method in combination with residual
connections. Since HouseGNN does not rely on additional stabilization combining it with
hyperparameters and residual connections may further improve stability and could
potentially lead to significantly higher accuracy.
\appendices
\section{Algorithm Summary}

Algorithm~\ref{alg:housegnn} summarizes the HouseGNN forward pass.

\begin{algorithm}[h]
\caption{HouseGNN Forward Pass}
\label{alg:housegnn}
\begin{algorithmic}[1]
\REQUIRE Graph $\mathcal{G}$ with edge set $\mathcal{E}$, features $X\in\mathbb{R}^{n\times d_{\mathrm{in}}}$,
         parameters $\{W^{(\ell)}\}_{\ell=0}^{L-1}$, $W_{\mathrm{enc}}$, $W_{\mathrm{cls}}$
\ENSURE Class logits $\{\ell_i\}_{i=1}^n$
\STATE $H^{(0)} \leftarrow XW_{\mathrm{enc}}^\top + \mathbf{1}\mathbf{b}_{\mathrm{enc}}^\top$
       \COMMENT{Encoder}
\FOR{$\ell = 0$ \TO $L-1$}
  \STATE $M^{(\ell)} \leftarrow \text{MeanAggr}(H^{(\ell)}, \mathcal{E})$
         \COMMENT{\eqref{eq:aggregate}}
  \STATE $V^{(\ell)} \leftarrow M^{(\ell)} (W^{(\ell)})^\top$
  \FOR{each node $i$}
    \IF{$\|\mathbf{v}^{(\ell)}_i\|_2 = 0$}
      \STATE $\mathbf{z}^{(\ell)}_i \leftarrow \mathbf{h}^{(\ell)}_i$
    \ELSE
      \STATE $\mathbf{u}^{(\ell)}_i \leftarrow \mathbf{v}^{(\ell)}_i / \|\mathbf{v}^{(\ell)}_i\|_2$
             \COMMENT{Node-wise normalisation; \eqref{eq:normalise}}
      \STATE $\mathbf{z}^{(\ell)}_i \leftarrow \mathbf{h}^{(\ell)}_i - 2\,\mathbf{u}^{(\ell)}_i(\mathbf{u}^{(\ell)\top}_i\mathbf{h}^{(\ell)}_i)$
             \COMMENT{Householder reflection; \eqref{eq:reflect}}
    \ENDIF
    \STATE $\mathbf{h}^{(\ell+1)}_i \leftarrow \text{GroupSort}(\mathbf{z}^{(\ell)}_i)$
           \COMMENT{\eqref{eq:groupsort}}
  \ENDFOR
\ENDFOR
\FOR{each node $i$}
  \STATE $\ell_i \leftarrow \mathbf{h}^{(L)}_i W_{\mathrm{cls}}^\top + \mathbf{b}_{\mathrm{cls}}$
         \COMMENT{Classifier}
\ENDFOR
\RETURN $\{\ell_i\}_{i=1}^n$
\end{algorithmic}
\end{algorithm}


\begin{thebibliography}{99}

\bibitem{kipf2017semi}
T.~N.~Kipf and M.~Welling,
``Semi-supervised classification with graph convolutional networks,''
in \textit{Proc.\ Int.\ Conf.\ Learning Representations (ICLR)}, 2017.

\bibitem{velickovic2018gat}
P.~Veli\v{c}kovi\'{c}, G.~Cucurull, A.~Casanova, A.~Romero, P.~Li\`o, and Y.~Bengio,
``Graph attention networks,''
in \textit{Proc.\ Int.\ Conf.\ Learning Representations (ICLR)}, 2018.

\bibitem{xu2019how}
K.~Xu, W.~Hu, J.~Leskovec, and S.~Jegelka,
``How powerful are graph neural networks?''
in \textit{Proc.\ Int.\ Conf.\ Learning Representations (ICLR)}, 2019.

\bibitem{xu2018representation}
K.~Xu, C.~Li, Y.~Tian, T.~Sonobe, K.~Kawarabayashi, and S.~Jegelka,
``Representation learning on graphs with jumping knowledge networks,''
in \textit{Proc.\ Int.\ Conf.\ Machine Learning (ICML)}, 2018.

\bibitem{wu2019simplifying}
F.~Wu, A.~Souza, T.~Zhang, C.~Fifty, T.~Yu, and K.~Weinberger,
``Simplifying graph convolutional networks,''
in \textit{Proc.\ Int.\ Conf.\ Machine Learning (ICML)}, 2019.

\bibitem{gilmer2017neural}
J. Gilmer, S. S. Schoenholz, P. F. Riley, O. Vinyals, and G. E. Dahl, “Neural message passing for quantum chemistry,” in Proc. Int. Conf. Machine Learning (ICML), 2017.

\bibitem{duvenaud2015convolutional}
D.~K.~Duvenaud \textit{et~al.},
``Convolutional networks on graphs for learning molecular fingerprints,''
in \textit{Proc.\ Neural Inf.\ Process.\ Syst.\ (NeurIPS)}, 2015.

\bibitem{li2018deeper}
Q.~Li, Z.~Han, and X.-M.~Wu,
``Deeper insights into graph convolutional networks for semi-supervised learning,''
in \textit{Proc.\ AAAI Conf.\ Artif.\ Intell.}, 2018.

\bibitem{oono2020graph}
K.~Oono and T.~Suzuki,
``Graph neural networks exponentially lose expressive power for node classification,''
in \textit{Proc.\ Int.\ Conf.\ Learning Representations (ICLR)}, 2020.

\bibitem{wu2023nonasymptotic}
X.~Wu, Z.~Chen, W.~Wang, and A.~Jadbabaie,
``A non-asymptotic analysis of oversmoothing in graph neural networks,''
in \textit{Proc.\ Int.\ Conf.\ Learning Representations (ICLR)}, 2023.

\bibitem{wu2023demystifying}
X.~Wu, A.~Ajorlou, Z.~Wu, and A.~Jadbabaie,
``Demystifying oversmoothing in attention-based graph neural networks,''
in \textit{Proc.\ Neural Inf.\ Process.\ Syst.\ (NeurIPS)}, 2023.

\bibitem{zhao2020pairnorm}
L.~Zhao and L.~Akoglu,
``PairNorm: Tackling oversmoothing in GNNs,''
in \textit{Proc.\ Int.\ Conf.\ Learning Representations (ICLR)}, 2020.

\bibitem{chen2020simple}
M.~Chen, Z.~Wei, Z.~Huang, B.~Ding, and Y.~Li,
``Simple and deep graph convolutional networks,''
in \textit{Proc.\ Int.\ Conf.\ Machine Learning (ICML)}, 2020.

\bibitem{rong2020dropedge}
Y.~Rong, W.~Huang, T.~Xu, and J.~Huang,
``Dropedge: Towards deep graph convolutional networks on node classification,''
in \textit{Proc.\ Int.\ Conf.\ Learning Representations (ICLR)}, 2020.

\bibitem{gasteiger2019predict}
J.~Gasteiger, A.~Bojchevski, and S.~Günnemann,
``Predict then propagate: Graph neural networks meet personalized PageRank,''
in \textit{Proc.\ Int.\ Conf.\ Learning Representations (ICLR)}, 2019.

\bibitem{liu2020dagnn}
M.~Liu, H.~Gao, and S.~Ji,
``Towards deeper graph neural networks,''
in \textit{Proc.\ ACM SIGKDD Int.\ Conf.\ Knowledge Discovery and Data Mining}, 2020.

\bibitem{scholkemper2025residual}
M.~Scholkemper, X.~Wu, A.~Jadbabaie, and M.~T.~Schaub,
``Residual connections and normalization can provably prevent oversmoothing in GNNs,''
in \textit{Proc.\ Int.\ Conf.\ Learning Representations (ICLR)}, 2025.

\bibitem{wang2025signed}
J.~Wang, X.~Wu, J.~Cheng, and Y.~Wang,
``A signed graph approach to understanding and mitigating oversmoothing in GNNs,''
\textit{arXiv:2502.11394}, 2025.

\bibitem{guo2021orthogonal}
K.~Guo, K.~Zhou, X.~Hu, Y.~Li, Y.~Chang, and X.~Wang,
``Orthogonal graph neural networks,''
\textit{arXiv:2109.11338}, 2021.

\bibitem{kiani2024unitary}
[20] B. T. Kiani, L. Fesser, and M. Weber, “Unitary convolutions for learning on graphs and groups,” arXiv:2410.05499, 2024.

\bibitem{qiu2024unitary}
H.~Qiu, Y.~Bian, and Q.~Yao,
``Graph unitary message passing,''
\textit{arXiv:2403.11199}, 2024.

\bibitem{anil2019sorting}
C.~Anil, J.~Lucas, and R.~Grosse,
``Sorting out Lipschitz function approximation,''
in \textit{Proc.\ Int.\ Conf.\ Machine Learning (ICML)}, 2019.

\bibitem{householder1958unitary}
A.~S.~Householder,
``Unitary triangularization of a nonsymmetric matrix,''
\textit{J.\ ACM}, vol.~5, no.~4, pp.~339--342, 1958.

\bibitem{mhammedi2017efficient}
Z.~Mhammedi, A.~Hellicar, A.~Rahman, and J.~Bailey,
``Efficient orthogonal parametrisation of recurrent neural networks using Householder reflections,''
in \textit{Proc.\ Int.\ Conf.\ Machine Learning (ICML)}, 2017.

\bibitem{he2016deep}
K.~He, X.~Zhang, S.~Ren, and J.~Sun,
``Deep residual learning for image recognition,''
in \textit{Proc.\ IEEE Conf.\ Comput.\ Vis.\ Pattern Recognit.\ (CVPR)}, 2016.

\bibitem{ioffe2015batch}
S.~Ioffe and C.~Szegedy,
``Batch normalization: Accelerating deep network training by reducing internal
covariate shift,''
in \textit{Proc.\ Int.\ Conf.\ Machine Learning (ICML)}, 2015.

\bibitem{mccallum2000automating}
A.~K.~McCallum, K.~Nigam, J.~Rennie, and K.~Seymore,
``Automating the construction of internet portals with machine learning,''
\textit{Inf.\ Retrieval}, vol.~3, no.~2, pp.~127--163, 2000.

\bibitem{giles1998citeseer}
C.~L.~Giles, K.~D.~Bollacker, and S.~Lawrence,
``CiteSeer: An automatic citation indexing system,''
in \textit{Proc.\ 3rd ACM Conf.\ Digital Libraries}, 1998.

\bibitem{namata2012query}
G.~Namata, B.~London, L.~Getoor, and B.~Huang,
``Query-driven active surveying for collective classification,''
in \textit{Workshop Mining and Learning with Graphs (MLG)}, 2012.

\bibitem{eliasof2021pde}
M.~Eliasof, E.~Haber, and E.~Treister,
``PDE-GCN: Novel architectures for graph neural networks motivated by partial
differential equations,''
in \textit{Proc.\ Neural Inf.\ Process.\ Syst.\ (NeurIPS)}, 2021.

\bibitem{rusch2023survey}
T.~K.~Rusch, M.~M.~Bronstein, and S.~Mishra,
``A survey on oversmoothing in graph neural networks,''
\textit{arXiv:2303.10993}, 2023.

\bibitem{shi2019dynamics}
G.~Shi, C.~Altafini, and J.~S.~Baras,
``Dynamics over signed networks,''
\textit{SIAM Review}, vol.~61, no.~2, pp.~229--257, 2019.

\bibitem{platonov2023critical}
O.~Platonov, D.~Kuznedelev, M.~Diskin, A.~Babenko, and L.~Prokhorenkova,
``A critical look at the evaluation of GNNs under heterophily: Are we really
making progress?''
\textit{arXiv:2302.11640}, 2023.

\end{thebibliography}
\end{document}